\documentclass[10pt,conference]{IEEEtran}
\usepackage[utf8]{inputenc}
\usepackage{graphicx}
\usepackage{caption}
\usepackage{subcaption}
\usepackage{amsmath}
\usepackage{color,colortbl}
\usepackage{xcolor}
\usepackage{soul}
\usepackage{siunitx}
\usepackage{units}
\usepackage{booktabs}
\usepackage{enumitem}
\usepackage{multirow}
\usepackage{amssymb}
\usepackage{pifont}
\usepackage[utf8]{inputenc}
\usepackage{xspace}
\usepackage{hyperref}
\usepackage{balance}
\usepackage{cleveref}
\usepackage{listings}
\usepackage{tabularx}
\usepackage[utf8]{inputenc}
\usepackage{fancyhdr} 
\usepackage{lipsum}
\usepackage{lmodern}
\usepackage{tcolorbox}
\usepackage{tabu}
\usepackage[numbers,sort&compress]{natbib}

\def\BibTeX{{\rm B\kern-.05em{\sc i\kern-.025em b}\kern-.08em
    T\kern-.1667em\lower.7ex\hbox{E}\kern-.125emX}}

\newcommand{\cmark}{\ding{51}}%
\newcommand{\xmark}{\ding{55}}%
\newcommand{\plm}{PLM\xspace}
\newcommand{\plms}{PLMs\xspace}

\newcommand{\zijian}[1]{{{}}}
\newcommand{\mkr}[1]{{{}}}
\newcommand{\rayb}[1]{{{}}}
\newcommand{\shiqi}[1]{{{}}}
\newcommand{\sujan}[1]{{{}}}
\newcommand{\wasi}[1]{{{}}}

\newcounter{RQACounter}

\newcommand{\RS}[2]{%
\begin{tcolorbox}[width=\columnwidth]
\vspace{-1mm}
\textbf{Result {#1}:~}{\emph {#2}}%
\vspace{-1mm}
\end{tcolorbox}
}

\title{Greener yet Powerful: Taming Large Code Generation Models with Quantization}
\DeclareRobustCommand*{\IEEEauthorrefmark}[1]{%
  \raisebox{0pt}[0pt][0pt]{\textsuperscript{\footnotesize *}}%
}
\author{
    \IEEEauthorblockN{Xiaokai Wei, Sujan Gonugondla, Wasi Ahmad, Shiqi Wang, Baishakhi Ray, \\ Haifeng Qian, Xiaopeng Li, Varun Kumar, Zijian Wang, Yuchen Tian, Qing Sun, \\ Ben Athiwaratkun, Mingyue Shang, Murali Krishna Ramanathan, Parminder Bhatia, Bing Xiang}
    \IEEEauthorblockA{AWS AI Labs
    \\\{xiaokaiw, gsujan, wuahmad, wshiqi, rabaisha, qianhf, xiaopel, kuvrun, zijwan,  tiayuche, \\ qinsun,  benathi, myshang, mkraman, parmib, bxiang\}@amazon.com}
}

\begin{document}
\bstctlcite{IEEEexample:BSTcontrol}

\maketitle

\linepenalty=1000
\begin{abstract}
ML-powered code generation aims to assist developers to write code in a more productive manner, by intelligently generating code blocks based on natural language prompts\mkr{ and existing code}.
Recently, large pretrained deep learning models have \mkr{delete substantially to avoid unquantifiable and avoidable adjectives} substantially pushed the boundary of code generation and achieved impressive performance. Despite their \mkr{rephrase} great power, the huge number of model parameters poses a significant threat to adapting them in a regular software development environment, where a developer might use a standard laptop or mid-size server to develop her code. Such large models incur significant resource usage (in terms of memory, latency, and dollars) as well as carbon footprint.

Model compression is a promising approach to address these challenges.  Several techniques are proposed to compress large pretrained models typically used for vision or textual data. Out of many available compression techniques, we identified that quantization is mostly applicable for code generation task as it does not require significant retraining cost. As quantization represents model parameters with lower-bit integer (e.g., \texttt{int8}), the model size and runtime latency would both benefit from such \texttt{int} representation. We extensively study the impact of quantized model on code generation tasks across different dimension: (i) resource usage and carbon footprint, (ii) accuracy, and (iii) robustness. To this end, through systematic experiments we find a recipe of quantization technique that could run even a $6$B model in a regular laptop without significant accuracy or robustness degradation. We further found the recipe is readily applicable to code summarization task as well.

\end{abstract}

\section{Introduction}
\label{sec:intro}

\begin{figure*}[!htpb]
    \centering
    \includegraphics[width=0.95\textwidth]{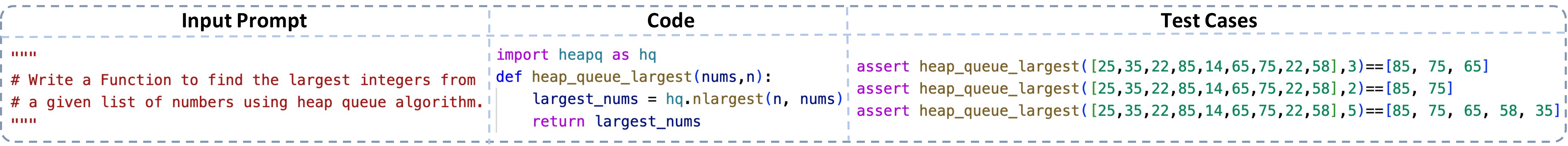}
    \caption{\small{\textbf{Sample prompt, code, and test cases taken from MBPP dataset~\cite{AustinJ2021}}. Given the NL prompt, a code generation model aims to generate the corresponding code. The associated test cases run the generated code to check functional correctness.}}
    \label{fig:example}
    \vspace{-10pt}
\end{figure*}

In recent years, ML-powered code generation tools, like Codex~\cite{chen2021evaluating}, GitHub Copilot~\cite{copilot}, Amazon CodeWhisperer\footnote{https://aws.amazon.com/codewhisperer/}, have gained significant traction.
These services aim to generate a computer program in response to a human-written specification (commonly called \textit{prompt}), as shown in~\Cref{fig:example}. Such tools bring promise to significantly automate the software development process and thus, improve developers' productivity. 

The backbone of ML-powered code generation tools are transformer based large pretrained language model (\plm)~\cite{AustinJ2021,ChowdheryA2022,NijkampE2022}. 
The Code Generation greatly benefits from the rapid development of \plms, as 
they have recently exhibited superior performance in multiple code-related tasks, including code generation, code summarization and type inference~\cite{ChenM2021, FengZ2020,LeH2022,FriedD2022,AustinJ2021,ChowdheryA2022,NijkampE2022}. 
Despite the great success, there are multiple challenges and downsides associated with applying the gigantic code generation models (2B-16B parameters) in a regular development environment.

\begin{itemize}[leftmargin=*]
    \item {\bf Hosting.} The huge number of model parameters poses a significant challenge. 
    For example, one of the largest open source models, CodeGen~\cite{NijkampE2022}, contains up to $16B$ parameters. Mere hosting this model in a regular desktop environment with a regular laptop becomes almost impossible, as it requires 72 GB of memory. A regular development laptop rarely comes with this much memory (A decent MAC laptop usually has 16 GB or 32 GB RAM). 
    Even if someone uses paid servers like EC2, using such models becomes extremely expensive\textemdash it might require around \$$100+$ per $1$k queries. Furthermore, their sizes will continue to grow, and accordingly, more stringent requirements and costs for hosting.  
    \item {\bf Latency and user experience.} The state-of-the-art code generation typically consists of $20 \sim 50$ transformer layers and $2B \sim 16B$ parameters. Model inference/serving on single GPU machine might incur a latency of several seconds. Such a delay in response would cause a negative user experience, especially for interactive code development.
    \item {\bf Carbon footprint.} Recently, researchers \cite{HendersonP2020}\cite{PattersonD2021} start to pay more attention to examining PLMs from the perspective of responsible and green AI. The training and inference of large PLMs typically involve a considerable amount of $\mathbf{CO_2}$ emission. For example,  the  $\mathbf{CO_2}$ emission of training GPT-$3$ model ($175$B parameters) amounts to three times that of a whole jet plane for San Francisco$\leftrightarrow$New York~\cite{PattersonD2021}.
\end{itemize}

To address these challenges, Machine Learning researchers started investigating different model compression techniques~\cite{schwartz2020green}. 
A key challenge, however, is to still preserve the powerfulness of the gigantic models  while significantly reducing the computational cost by compressing them.  Addressing this challenge would be crucial to democratizing the power of AI. 
In this paper, we empirically investigate whether such model compression techniques can be effective for code generation models.

Our target user is a regular developer using a laptop with a good configuration (e.g., a Laptop with CPU only/limited GPUs or with access to a moderate sized server). She uses a state-of-the-art code generation model. Also, she does not have resources to retrain a huge PLM from scratch.  In such a scenario, we identify the following desirable properties that a practically useful model compression strategy needs to satisfy:

 \begin{itemize}[leftmargin=*]
 \item {\bf Minimal compression cost}: converting a pretrained model to a more efficient version typically involves certain processing/training costs. If the compression technique requires significant (re)training of the large model over substantial amounts of data, it could result in undesirable environmental impacts (large power consumption and carbon footprint) and the cost would be prohibitively high for an average user to afford. High processing costs would contradict the purpose of greener AI and democratizing AI.
 \item {\bf Substantial reduction in hosting cost:} as state-of-the-art models are already gigantic (e.g., $2$B to $16$B parameters) and are expected to continue growing in sizes, minor reductions in compressed size or runtime latency would not be practically useful. Ideally, one would expect a properly designed model compression method to bring at least $50$\% improvement in these key hosting metrics (e.g., size/latency).
 \item {\bf Preservation of generation power}: it is highly desirable the compressed model would still have similar generation power as the original model. Model compression at the cost of significantly degenerated predictions would make the compressed model much less appealing to employ.
 \item {\bf Minimal adverse side effect}: in addition to preserving generation accuracy, we also expect the model to not degenerate in other important aspects of generation, such as weakened robustness.
 \end{itemize}
 

Most model compression techniques developed by ML community, such as distillation \cite{WangF2021,SunZ2020}, pruning \cite{JiaoX2020,LagunasF2021} and quantization-aware training \cite{ZhangW2020,TaoC2022,li2022dq} are often associated with large training costs. Training or finetuning large transformer models requires access to training data and large compute resources. This is often not an option for many users who typically use the model pretrained on large training corpus by others.

Out of many model compression options, we are able to identify a compression recipe with negligible processing cost and preserved accuracy with a specific subcategory of quantization methods, i.e., Post-Training Quantization (PTQ). Quantization is a compression technique where the weights and activations of an ML model are converted to and computed with integer data types such as \texttt{int8} instead of commonly used float-point data types such as \texttt{fp32}. As data is represented with lower-bits (e.g., $8$ or $4$) the model would be much smaller in size. Also, most hardware types (either CPU or GPU) perform integer operations (e.g., multiplication) at a much faster speed; the quantized model would also likely to enjoy reduced computational cost and latency. Properly designed PTQ methods would require none or a relatively small amount of code data for post-training processing, and experimental results show that the proposed approach is highly effective on multiple tasks. This means one can get all the compression benefits (e.g., latency/memory/storage/carbon emission) with negligible cost while retaining the generation power of the full-precision model.

Our contribution can be summarized as follows:
\begin{itemize}[leftmargin=*]
    \item  We recognize the importance of model compression in the context of code generation and identify the adequacy of post-training quantization for this purpose. To our best knowledge, this is the first attempt at compressing  a state-of-the-art code generation model. Impact-wise, the quantized model with $16$B parameters could run on a personal laptop with only CPUs, and generate a $20$-token long prediction within $25$ seconds (as opposed to $70$ seconds by the corresponding full-precision model).
    \sujan{This 2 second claim is not true for a 2B model, it was for a 350M model}
    \item We perform an extensive empirical study on multiple code generation models with their quantized variations on both NL-to-code and code-to-NL tasks. We observe comparable accuracy across multiple types of models and parameter sizes with the proposed quantization techniques. Even for extremely large CodeGen-$16B$, we can preserve comparable accuracy with quantization. Besides, we experiment in different ablation settings to provide guidelines for properly employing quantization.
    \item  We present an in-depth empirical analysis on the layers, activations, and weights of the  state-of-the-art code generation models to gain deeper insights on the effect of quantization in them. This helps us understand why certain quantization methods perform better than others.
    \item Beyond accuracy, we also investigate the impact of quantization on model robustness, which is often overlooked
    by the existing code generation literature. We show that the proposed quantization recipe would have no adverse impact on model robustness.
\end{itemize}


\section{Background \& Related Work}
\label{section:background}

\subsection{Code Generation with Transformer-based Models}


Recently, applying transformer-based Pretrained Language Models (PLMs) to the source code generation task, have drawn considerable attention and set overwhelmingly strong state-of-the-art in this field ~\cite{AustinJ2021,ChowdheryA2022,NijkampE2022,ChenM2021,LeH2022,FriedD2022}. 
The goal is to generate complete or code fragments given natural language or partial code as prompts.
To achieve this goal, large language models are trained on humongous code corpora, typically curated from open source code archives like GitHub, Stack Overflow, etc.

\begin{figure}
     \centering
     \includegraphics[width=0.48\textwidth]{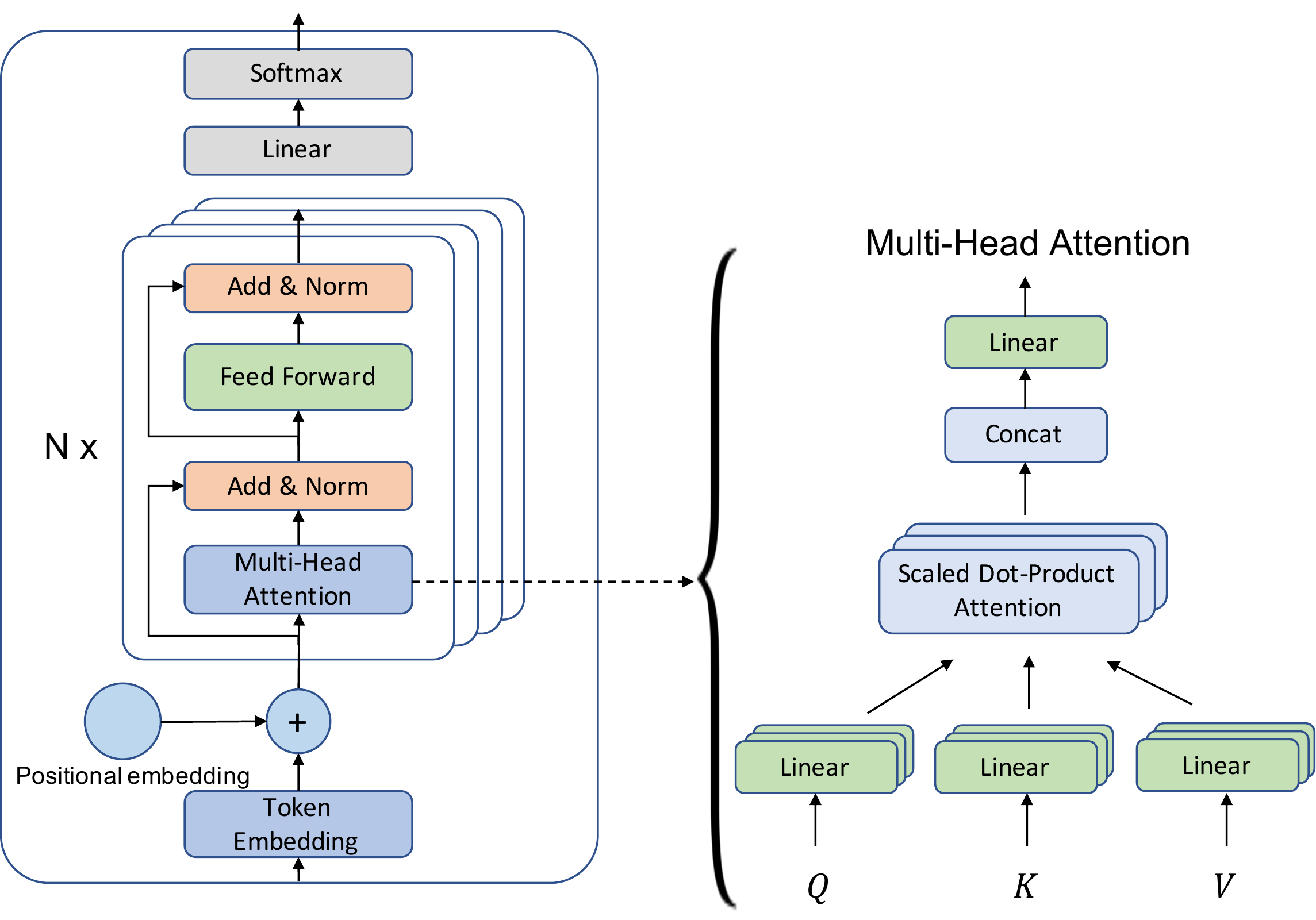}
     \caption{\small{\textbf{Transformer structure and multi-head attention cell. The feed-forward layer and all linear layers inside multi-head attention are colored in green. We quantize all these linear layers in the network.}}}
     \label{fig:transformer}
\end{figure}



The PLMs typically use decoder-only (e.g., GPT~\cite{RadfordA2019}) or encoder-decoder architecture (e.g., BART~\cite{LewisM2020}/T5~\cite{RaffelC2020}). 
For code generation tasks, decoder-only models (e.g., CodeGen~\cite{NijkampE2022} and Incoder~\cite{FriedD2022})  take some pre-encoded code representation and learn to decode, i.e., synthesize next token sequences. 
Typically, these models use causal language modeling, i.e, generate the tokens conditioned on the previous token sequences. Thus, decoder-only models are a natural fit for code completion tasks where the previous code context is given and the model is expected to generate the next tokens.
In contrast, encoder-decoder based code generation models like PLBART~\cite{AhmadW2021} and CodeT5~\cite{WangY2021} are typically trained to learn to reconstruct the original code sequence that is corrupted using an arbitrary noise function. Therefore, such models do not naturally fit the code completion tasks but are found effective when finetuned for code generation or summarization tasks. 



\subsection{Model Compression}
The large transformer models use billions of parameters and may require trillions of operations for generating code. Model compression tackles this high costs of large models to enable their wider and easier adoption. Model compression is a class of techniques designed to reduce model size (i.e., bytes required to represent the model) and improve generation latency while maintaining minimum accuracy (i.e., ability to generate useful and correct code) degradation. Some representative techniques include:

\begin{enumerate}
    \item \textit{Knowledge distillation.} A small student model is trained on the outputs of a larger teacher model that we want to compress~\cite{WangF2021,SunZ2020}.
    \item \textit{Pruning.} It constitutes  a class of techniques that make the weight matrices\mkr{define a weight matrix} sparse to reduce the number of parameters as many of the matrix entries will now be zeros \cite{JiaoX2020,LagunasF2021,WangZ2020,XiaM2022}.
    \item \textit{Quantization.} This technique uses fewer bits to represent the weights of parameterized functions \cite{BondarenkoY2021,ZhangW2020}.
\end{enumerate}




\begin{figure}[t]
\centering
\includegraphics[width=0.48\textwidth]{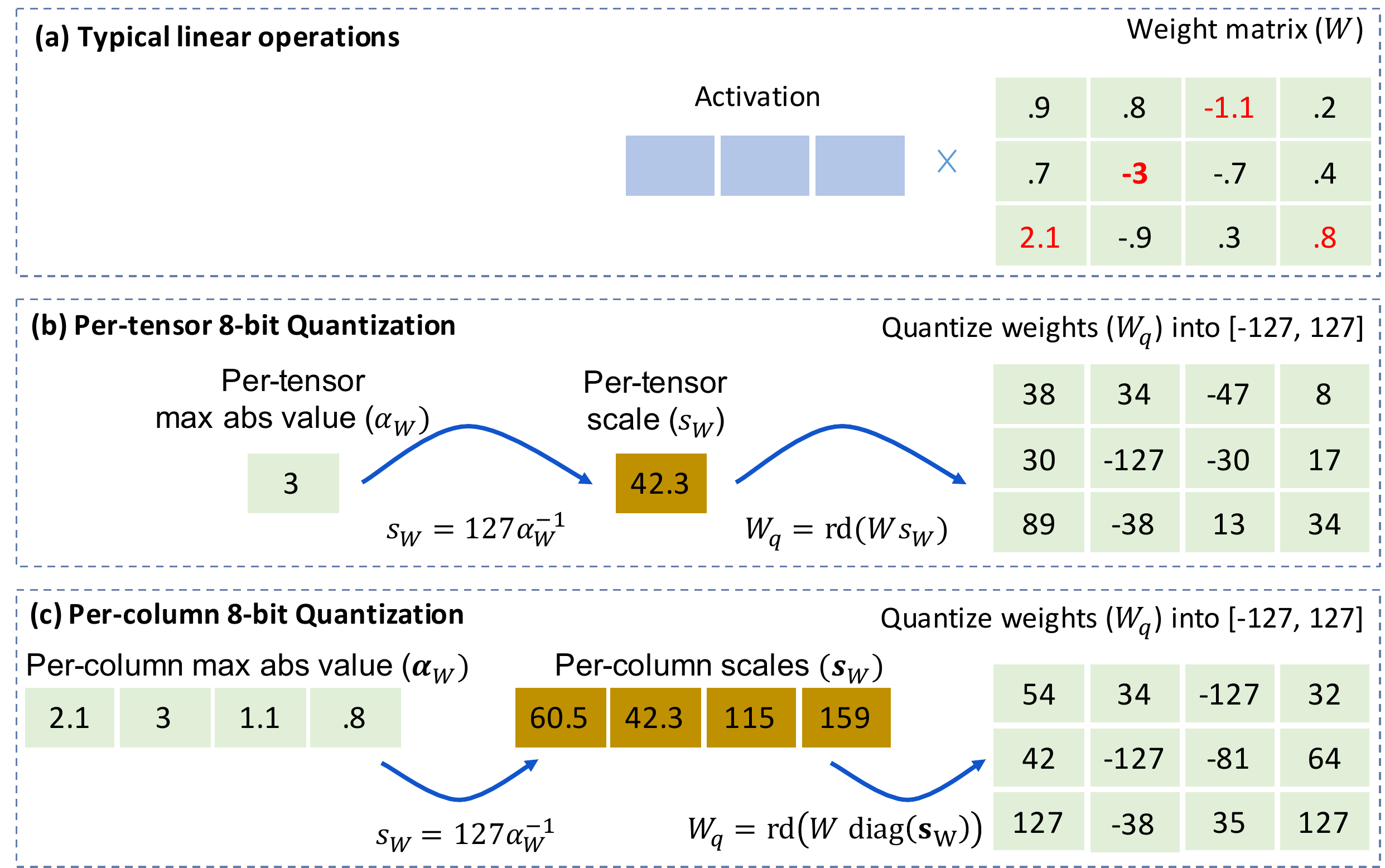}
 \caption{\textbf{\small{Toy example for quantizing the typical floating-point weight matrix (a) into \texttt{int8} matrix using (b) per-tensor v.s. (c) per-column quantization.}}}
 \label{fig:pertensor}
\end{figure}

\subsection{Quantization for model compression}

Here we describe the process of quantizing a tensor and discuss different 
model-quantization techniques. 


\subsubsection{Quantization operation}
\label{sec:QuantOpp}
Quantization refers to the conversion of a full-precision (or floating-point) tensors to tensors with integer values. An example of the quantization operations is depicted in Figure~\ref{fig:pertensor}. Given a matrix $W$, a basic quantizer $Q(\cdot)$ uses scale and rounding operations to get the quantized version of the matrix:
\begin{equation*}
    Q(W) = \frac{W_q}{s_W},\ \text{where}\ s_W=\frac{2^{B-1}}{\alpha_W}\ \text{and}\ W_q= round(s_W W)
\end{equation*}


\noindent Here, $\alpha_W$ is the quantization
range, and $B$ is the bitwidth (which is 8 in case of \texttt{int8}), $W_q$ is the quantized integer matrix, $s_W$ is the quantization scale, and $Q(W)$ is the quantized approximation of the matrix $W$

\noindent
\textbf{Quantization Noise.} We assess the quality of quantization by estimating the relative quantization noise $q_a$, defined as \cite{charbel}:
\begin{align}
    q_a = \frac{||A-Q(A)||_2}{||A||_2} \approx \frac{\Delta_A^2}{12||A||_2} \approx \frac{1}{12 s^2_A||A||_2}
\end{align}
where $||x||_2$ is the the $L_2$-norm of the vector $x$, and $\Delta_W = 1/s_W$ quantization step size. The quantization noise increases with $\Delta_W$ (or decreases with $s_W$), as the approximation of the full precision parameters becomes coarser.

\begin{figure}[h]
     \centering
         \includegraphics[width=0.3\textwidth]{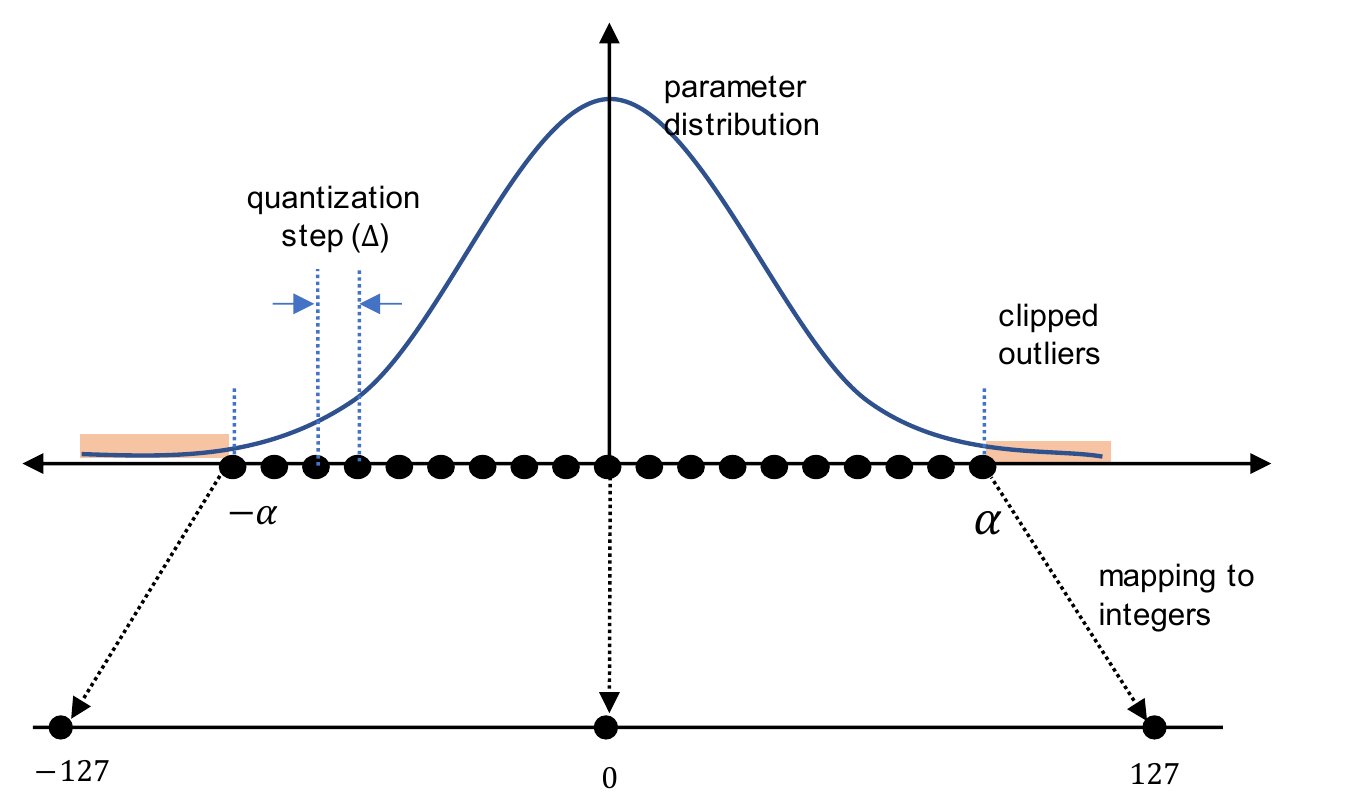}
         \caption{\small{\textbf{Illustration of quantization operation showing, quantization step, clipping, scaling, and mapping.}}}
         \label{fig:quantcartoon}
\end{figure}

\noindent
\textbf{Quantization Range and Scale Factor.} 
The quantization range $\alpha_W$ is the value that will be mapped to the largest representable integer (127 in the case of \texttt{int8}). Typically we set $\alpha_W = \mathrm{max}(\mathrm{abs}(W))$, consequently setting the scale factor $s_W=2^{B-1}/\mathrm{max}(\mathrm{abs}(W))$. However, having a large outlier in $W$ will increase $\alpha_W$ and therefore increase the quantization noise. To avoid that, some choose to clip the data by choosing $\alpha_W<\mathrm{max}(\mathrm{abs}(W))$ (see~\Cref{fig:quantcartoon}), where the matrix elements are clipped the $s$ before the quantization operation; i.e., matrix elements $>\alpha_W$ are set to $\alpha_W$ and those $<-\alpha_W$ are set to $-\alpha_W$. 


\subsubsection{Quantization techniques}
\label{sec:quant-technique}

Model quantization techniques can be classified based on the following:

\begin{figure}[htpb]
      \centering
     \begin{subfigure}{0.48\textwidth}
         \centering
          \includegraphics[width=\textwidth]{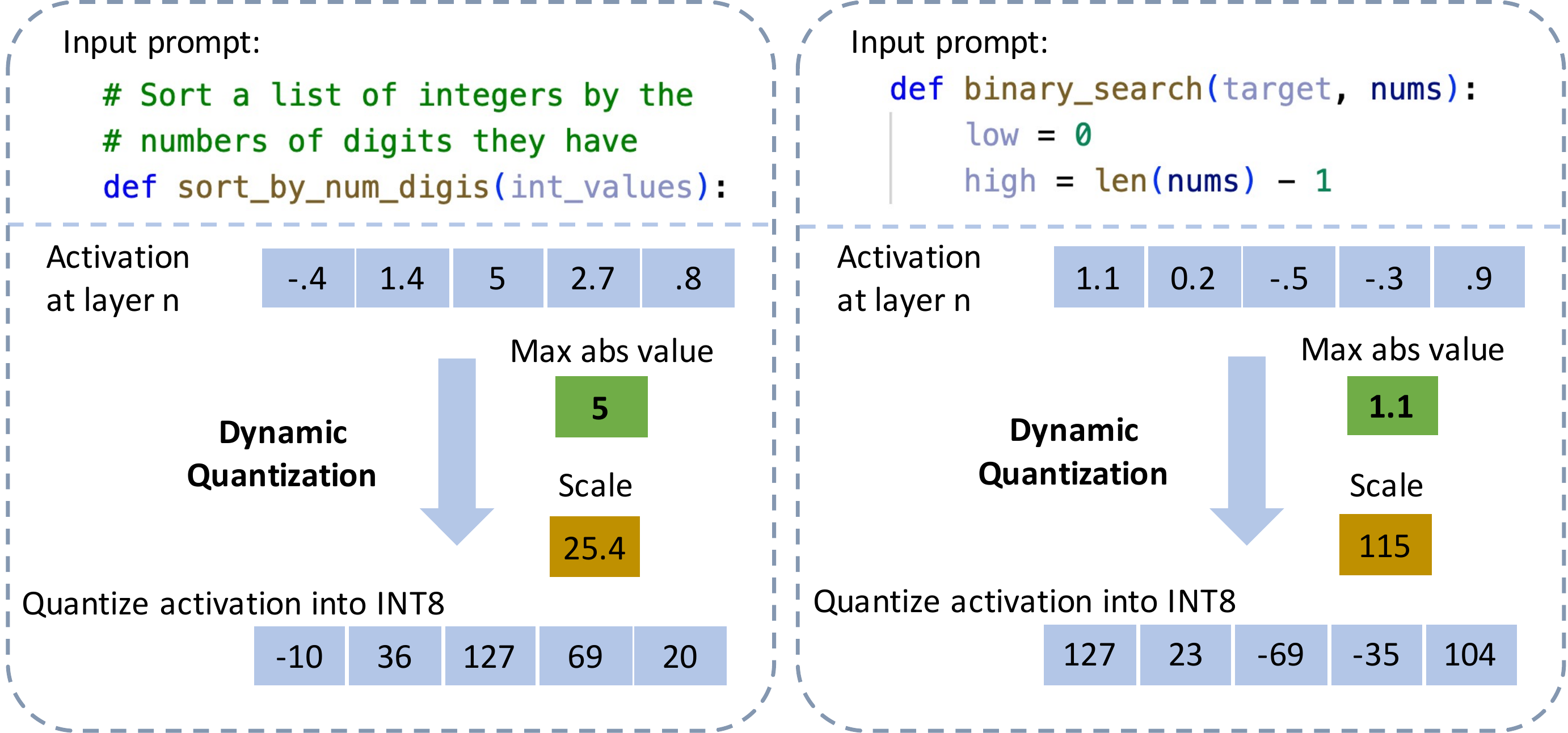}
          \caption{\small{\textbf{Dynamic quantization where the clipping range is determined dynamically as the max abs value in activation tensors.}}}
          \label{fig:decoder_model}
      \end{subfigure}
      \hfill
      \begin{subfigure}{0.48\textwidth}
          \centering
          \includegraphics[width=\textwidth]{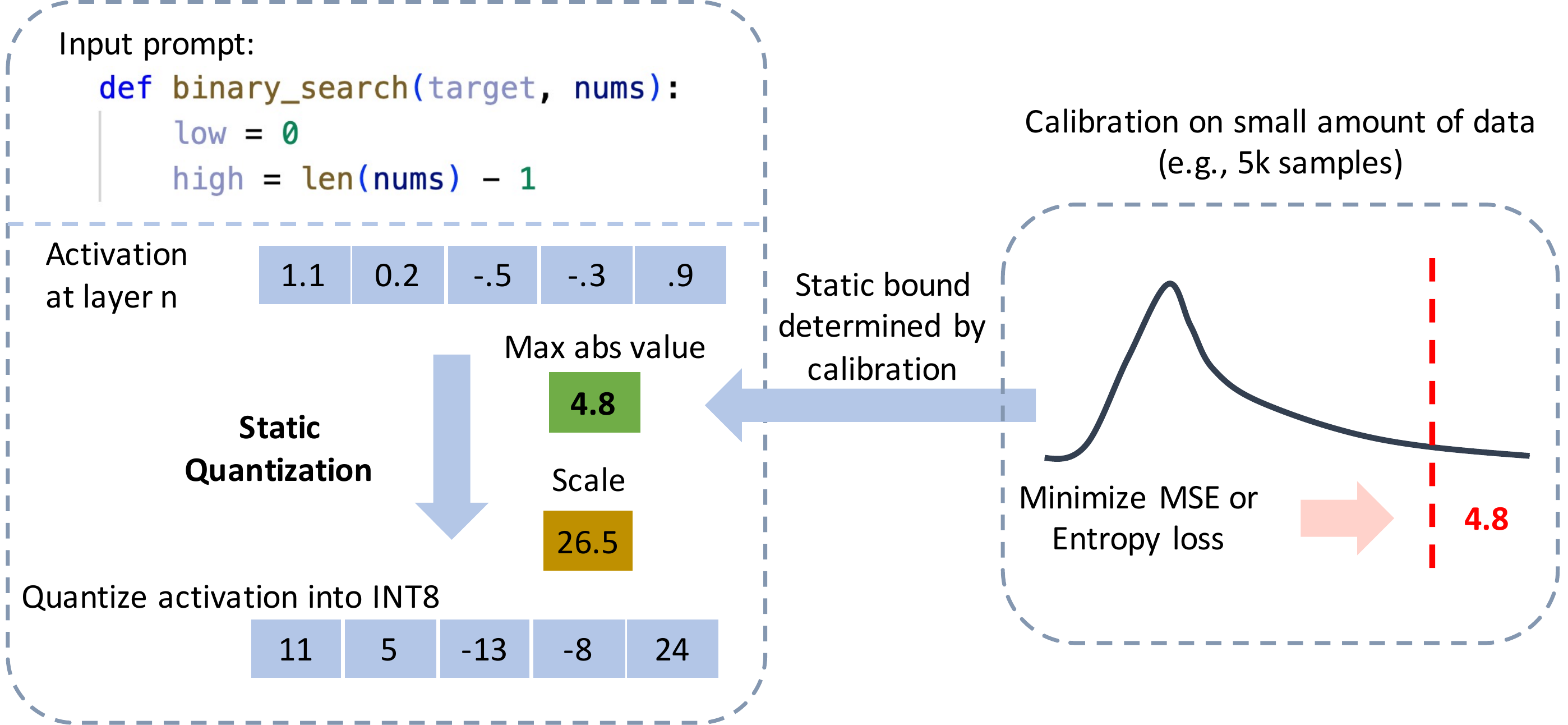}
          \caption{\small{\textbf{Static quantization where the fixed clipping range is learned through calibration.}}}
          \label{fig:transformers}
      \end{subfigure}
      \caption{\textbf{\small{Toy example on quantizing activations with dynamic quantization v.s. static quantization.}}}
      \label{fig:static-dynamic}
 \end{figure}
 
\noindent

\noindent
{ \bf Methods to obtain quantized network.} 
This can be broadly classified into

\begin{itemize}[leftmargin=*]
\item {\it Quantization Aware Training (QAT):}
QAT requires training the model from scratch with additional simulated quantization operations during training process to ensure the learned parameter values are quantization-friendly. This is expensive due to the potentially huge cost of training but would lead to models that potentially have higher accuracy than a PTQ model.
\item {\it Post Training Quantization (PTQ)}: PTQ derives a quantized (e.g., \texttt{int8}) network from an existing full-precision network without any training or finetuning with additional data. Since the model is not originally trained to perform inference with quantized parameters and activation, models quantized by PTQ tend to be more susceptible to quantization noise. However, the low costs associated with PTQ make it a very popular choice for obtaining quantized models.
\end{itemize}

\noindent
{\bf Methods to choose the activation scale.}
Here we choose the ranges and the values of the activations change for each example. There are various options to choose the quantization scale parameters for activations that can be classified into (see Figure~\ref{fig:static-dynamic}): 
\begin{itemize}
\item {\it Dynamic quantization:} Here we determine the clip range ($\alpha$) and scale parameter ($s$) on the fly for activations, in order to minimize quantization noise where possible. One could typically use the maximum (absolute) value of the activation tensors as the clip range for each input. However, determining the clip range dynamically would incur an additional scanning cost to find the max value. 
\item {\it Static quantization:} Here we use the same pre-determined scale through so-called \textbf{calibration} on samples by minimizing certain loss (e.g., MSE/Entropy) between original activation and quantized activations. Static quantization might be susceptible to higher quantization noise though it would lower computational cost during inference.
\end{itemize}

\noindent
{\bf Quantization Granularity.} 
As we discussed in the Section~\ref{sec:QuantOpp}, choosing a large clip range (accordingly, small scales $s_A$) due to outliers can lead to a large quantization step which adds to quantization noise. 
To avoid outliers, column-wise quantization scales can be used where the scales are selected based on the max value of each column instead of the entire matrix. Broadly, we can classify quantization techniques based on the granularity of the quantization scales into 1) {\it per-tensor scales} where the entire tensor uses a single scale $s_A$, and
2) {\it per-column/per-row scales} where each column uses a different scale. 

Figure~\ref{fig:pertensor} illustrates the differences between the two scaling options. Here, we are choosing scale values based on the maximum absolute value of the quantized block. Choosing per-column scales avoids tensor-wide outliers and allows for finer quantization steps than per-tensor scales.


In the rest of the paper, we will primarily use PTQ as this has minimal post/re-training cost.
We examine the accuracy of the models with dynamic and static quantization and discuss the impacts of choosing per-tensor and per-row scales. We will use \texttt{int8} precision for quantization as it is widely supported across all major CPUs and GPUs that are used today. 



\section{Methodology}
\label{sec:method}

\begin{table*}[htpb]
\centering
\caption{\textbf{\small{Model details under investigation.}}}
\label{table:model_to_study}
\setlength{\tabcolsep}{3pt}
\begin{tabular}{l l r r l l}
\toprule
\multirow{2}{*}{\bf Models} & \multirow{2}{*}{\bf \#Parameters} & \multicolumn{3}{c}{\bf Training Cost} & \multirow{2}{*}{\bf Architecture}  \\ 
\cmidrule{3-5}
& & {\bf \#Steps} / {\bf \#Epochs} & {\bf Data (approx.)} & {\bf Compute Resources} & \\
\midrule
PLBART  & 140M / 406M & 100k / - & 250 GiB & 8 NVIDIA
GeForce RTX 2080 Ti & Encoder-Decoder \\  
Code-T5  & 60M / 220M / 770M & - / 150 & 25 GiB & 16 NVIDIA A100 GPUs & Encoder-Decoder  \\  
InCoder  & 1.3B / 6.7B & - / 1 & 216 GiB & 248 NVIDIA V100 GPUs & Decoder-only \\ 
CodeGen  & 350M / 2B / 6B / 16B & 650k / - & 1812 GiB & Google's TPU-v4 & Decoder-only \\ 
\bottomrule
\end{tabular}
\vspace{-10pt}
\end{table*}


The goal of this work is to provide an empirical and conceptual analysis of 
quantization techniques, originally developed as a core ML technique, in the context of 
large code generation models. 
To do that, we analyze the characteristics of the models using different dimensions of quantization techniques, as discussed in~\Cref{sec:quant-technique}. This section discusses our study methodology in detail.

\subsection{Quantized Model Preparation}
\label{sec:prep}

\subsubsection{Schemes for Quantization} 

For quantization techniques, we investigate both schemes of quantization (dynamic and static) described in previous sections and prepare the quantized models as follows.
\begin{itemize}[leftmargin=*]
\item {\bf Dynamic quantization:} For implementation, we use the native PyTorch Quantization \footnote{https://pytorch.org/docs/stable/quantization.html} API and convert all the weight matrices in Feed Forward Network (FFN) and Self-Attention to \texttt{int8}. As explained in the previous section, the min/max bound of each layer's activation is determined in a dynamic manner depending on the input during inference. The processing time needed for this scheme is minimal, which typically takes $<1$ minutes for $2B$ models and $<4$ minutes for $6B$ models.
\item{\bf Static quantization:} Static quantization needs to determine the clipping range for activations before inference, and such ranges are typically obtained from calibration by minimizing the quantization noise. We perform the activation-bound calibration with a tiny fraction ($5$k samples) from the CodeSearchNet (Python) training set. In preliminary experiments, we find {\it MSE (Mean Squared Error)} based loss to be most effective, so we minimize the MSE between the quantized activations and full-precision ones as the calibration. 
\end{itemize}

\subsection{Study Subjects}
\label{sec:subj}

\noindent
\textbf{Studied Models.} We leverage the state-of-the-art and representative code generation models that have open sourced model checkpoints available, to study the efficacy of different calibration techniques. We aim to cover models with different sizes and backbone architectures. In particular, we focus on CodeGen \cite{NijkampE2022}, as they open sourced models with different sizes \{$350$M, $1$B, $6$B, $16$B\} and different language support (mono v.s. multi-language generation). Additionally, we also include InCoder \cite{FriedD2022}
to further confirm the patterns we observe with CodeGen models. 
We also studied two more models Code-T5 \cite{WangY2021} and PLBART \cite{AhmadW2021} for code summarization task.
The statistics of these models are summarized in~\Cref{table:model_to_study}.

\smallskip
\noindent
\textbf{Studied Tasks.}
In this paper, our main focus is code generation task (NL-to-code). 
Further, to stress test the effectiveness of quantization on other generative tasks, we 
study code summarization task for models' accuracy evaluation (RQ4). 
Thus, we study the following two tasks: 
\begin{itemize}[leftmargin=*]
    \item {\bf NL-to-code generation}: Here we evaluate the models' code generation ability.
    A user gives a natural language prompt as input. These are loosely defined specifications. The model is expected to generate the corresponding code fragments. The generated code is tested by running the test cases.~\Cref{fig:example} shows an example.

    \item {\bf Code-to-NL generation}: We further evaluate a generative model's capability on code summarization task, where given the function signature and body, the model generates an NL description of the function. 
\end{itemize}

\smallskip
\noindent
{\bf Studied Dataset:} We use HumanEval \cite{ChenM2021} and MBPP \cite{AustinJ2021} for evaluating the functional correctness of generated programs. 
The MBPP dataset~\cite{AustinJ2021} 
contains $974$ short Python functions with their textual descriptions and test cases to evaluate correctness (see~\Cref{fig:example}). 
HumanEval~\cite{ChenM2021} is a similar dataset released by OpenAI, which is  widely used in evaluating code generation tasks.  It contains $164$ hand-written Python programs, associated with their natural language descriptions and test cases. 

\smallskip
\noindent
{\bf Evaluation Metrics:} Generative models in NLP domain traditionally use some form of textual matching (exact or fuzzy match) between the generated text and ground truth and often report BLEU scores. Such textual similarity is problematic for evaluating code generation tasks, as the same functionality can be implemented in many ways. 
To overcome this, recent papers on code generation task~\cite{chen2021evaluating, kulal2019spoc, roziere2020unsupervised} recommend to evaluate functional correctness by running the generated code against test cases. Here we follow a similar evaluation criterion.

Each sample in our studied dataset 
is equipped with multiple test cases, as shown in~\Cref{fig:example}. The generated code needs to pass {\em all} provided tests to be considered as ``\textit{pass}''. 
Following ~\cite{kulal2019spoc,chen2021evaluating}, we report  pass@k to estimate the model's ability to generate code that will ``\textit{pass}''. 
Pass@$k$ measures the fraction of examples that are ``\textit{pass}'' by at least one of the $k$ solutions that the model generates.
However, given the ML model is probabilistic, we expect pass@$k$ to have a high variance. 
To address this, a standard practice is to generate $n>k$ solutions and estimate the statistical mean of pass@$k$ from these $n$ samples, i.e., estimate the fraction of times we ``\textit{pass}'' if we randomly pick $k$ samples from $n$.
In this paper, we use pass@$1$ and pass@$5$ as a metric for evaluations, which is estimated by generating $10$ samples per problem in the dataset. 
The reported accuracy (pass@$k$) is averaged on all samples generated for all programs in each dataset.


To evaluate the code summarization models, we use smoothed BLEU score \cite{lin-och-2004-orange} following prior works \cite{AhmadW2021, WangY2021}.

\section{Results} 
\label{section:experiment}

We evaluate the effect of quantization across three dimensions: greener, accuracy, and robustness for code generation tasks. To evaluate generalizability, we further evaluate quantization techniques for code summarization tasks, as code summarization is a popular code-related generative task where a different modality, i.e., text, is generated. In particular, we aim to answer the following four research questions: 
\begin{itemize}[leftmargin=*]
    \item {\bf RQ1.} How effective are quantization techniques for greener code generation models?
    \item {\bf RQ2.} Can quantized models maintain the prediction power of the corresponding full-precision models? 
    \item {\bf RQ3.} How robust are quantized models compared to the corresponding full-precision models?
    \item {\bf RQ4.} Are quantization techniques effective for other code related generative tasks such as code summarization?  
\end{itemize}


\subsection{Quantization for Greener Code Generation (RQ1)}

\noindent
\textit{Motivation.} The heart of this paper lies on this RQ, i.e., whether a quantized model can be substantially greener than its full precision counterpart. Here by green, we mean less resource usage and less carbon footprint. 
Our use case is to facilitate a regular development environment that can benefit from such large models. 
Thus, a full precision model can be pretrained with larger resource (even at industry scale).
However, a developer will be using the model in an environment which is either CPU-only or contain a smaller number of GPUs.
To this end, this RQ evaluates the model's resource usage and carbon footprint at inference time.

\noindent
\textit{Experimental Setup.} 
We aim to answer RQ$1$ by investigating quantization from a model hosting perspective, with GPU or CPU as the underlying hardware. 
We consider both on-cloud and on-device settings as both can be important use cases for code generation models. The environment used for experiment is the following:
\begin{itemize}[leftmargin=*]
    \item {\bf On cloud}: We use an AWS \texttt{p3dn.24xlarge} instance\footnote{More details on the hardware specification can be found at https://aws.amazon.com/ec2/instance-types/p3/} which have both CPUs and GPUs available with NVMe-based SSD storage.
    \item {\bf On device}: We use a typical developer's laptop, a MacBook Pro which runs macOS Monterey (version  $12.5$), with $32$ GB memory and M$1$ processor.
\end{itemize}

\noindent
\textit{Metrics.} 
We report inference latency and model storage size as primary metrics for model hosting. Based on the latency result and the specification of underlying hardware, we also estimate (assuming sequential prediction) the potential cost\footnote{Based on estimate in \resizebox{0.3\textwidth}{!}{https://www.instance-pricing.com/provider=aws-ec2/instance=p3dn.24xlarge}} (in US$\$$) and carbon emission \footnote{ \resizebox{0.4\textwidth}{!}{https://engineering.teads.com/sustainability/carbon-footprint-estimator-for-aws-instances/}} (in $gCO_2eq$) for evaluating the impact in terms of green AI.

\begin{table}[h]
\caption{\small{\textbf{Comparison on different hosting metrics between full-precision and quantized (\texttt{int8} dynamic) versions of Codegen-2B, Codegen-6B and Incoder-6B}.}}
\label{table:cpu_latency}
\centering
\setlength{\tabcolsep}{2pt}
\resizebox{0.48\textwidth}{!}{  
\begin{tabular}{l|ll|ll|ll}
\toprule
        & \multicolumn{2}{c|}{\textbf{Codegen-2B}} & \multicolumn{2}{c|}{\textbf{Codegen-6B}}  &  \multicolumn{2}{c}{\textbf{Incoder-6B}}  \\ \midrule
   \textbf{On Cloud / Precision}    &  \textbf{\texttt{fp32}}      & \textbf{\texttt{int8}}      &  \textbf{\texttt{fp32}}    & \textbf{\texttt{int8}}  & \textbf{\texttt{fp32}}    & \textbf{\texttt{int8}}   \\ \midrule
Storage (GB) &   $10.7$  &  $3.5$  &  $27.1$ &  $7.9$ & $25.4$ & $7.6$ \\ 
Latency (s/pred.) &  $3.47$  &  $2.47$  &    $7.81$    & $4.02$    &  $7.38$ &  $3.32$    \\ 
Est. $\mathrm{gCO}_2\mathrm{eq}$ ($1$k pred.) & $1309$   &  $932$ &   $2940$     &  $1516$  & $2783$  & $1252$  \\ 
Est. pricing ($1$k pred.) & \$30.1    & \$21.4   &  \$67.7   & \$34.8  &  \$64.0  &  \$28.8  \\ \midrule
    \textbf{ On Device}  
& & & & & & \\ \midrule
Latency (s/pred.) &    $23.7$  &  $10.7$ & $70.4$ & $25.4$ &  $52.6$ & $18$\\ 
\bottomrule
\end{tabular}
}
\end{table}

\begin{table}[h]
\centering
\caption{\small{\textbf{Latency on a GPU of a linear layer from the FFN network of CodeGen models in milli-seconds (ms) measured with CUTLASS kernels}.}}
\label{table:gpu_latency}
\setlength{\tabcolsep}{2pt}
\begin{tabular}{l|l|l|l} \toprule
                                     & \textbf{2B-FFN} & \textbf{6B-FNN} & \textbf{16B-FNN} \\ \midrule
{\bf Layer dims.} : &   $2560\rightarrow 10240$  &   $4096\rightarrow 16384$    &          $6144 \rightarrow 24576$ \\ \midrule 
\texttt{fp32}                                 &        $1.55\pm0.014$  &    $1.81\pm0.026$    &      $2.11\pm0.027$            \\
\texttt{fp16}                                 &       $1.10\pm0.011$           &   $1.28\pm0.012$    &  $1.68\pm0.022$   \\
\texttt{int8}                                 &       $0.90\pm0.016$          &    $1.08\pm0.014$ &     $1.29\pm0.11$            \\ \bottomrule
\end{tabular}
\end{table}

\begin{table*}[t]
\caption{\textbf{\small{Pass@k (\%) accuracy on HumanEval and MBPP.}} Performance gains are in blue and drops in red.}
\label{tab:rq2}

\centering
\setlength{\tabcolsep}{5pt}
\resizebox{\textwidth}{!}{  
\begin{tabu}{l|l|rr|rr|rr|rr}
\toprule 

& & \multicolumn{2}{c|}{\multirow{2}{*}{\textbf{Full-precision}}} & \multicolumn{2}{c|}{\textbf{Dynamic Quant.}} & \multicolumn{4}{c}{\textbf{Static Quant.}} \\ 
\cmidrule{5-10}
{\bf Dataset} & {\bf Model} & & & \multicolumn{2}{c|}{\textbf{(per-tensor)}} & \multicolumn{2}{c|}{\textbf{(per-tensor)}} & \multicolumn{2}{c}{\textbf{(per-column)}} \\
\cmidrule{3-10}
& & \textbf{pass@1} & \textbf{pass@5} & \textbf{pass@1}  & \textbf{pass@5}  & \textbf{pass@1}  & \textbf{pass@5}  & \textbf{pass@1}  & \textbf{pass@5}            \\ \midrule
\multirow{5}{*}{HumanEval} & Incoder-1.3B & 7.13 & 8.98 & 5.55 ({\color{red} \bf -1.58})  & 8.33 ({\color{red} \bf -0.65})   & 5.85 ({\color{red} \bf -1.28})  & 7.99 ({\color{red} \bf -0.99})  & 6.71 ({\color{red} \bf -0.42})  & 8.86 ({\color{red} \bf -0.12}) \\
& Incoder-6.7B          & 8.11          & 9.70         & 8.23 ({\color{blue} \bf +0.12})  & 10.52 ({\color{blue} \bf +0.82}) & 8.41 ({\color{blue} \bf +0.30})  & 10.46 ({\color{blue} \bf +0.76}) & 9.27 ({\color{blue} \bf +1.16})  & 11.38 ({\color{blue} \bf +1.68}) \\
& Codegen-350M          & 11.71         & 16.21         & 11.77 ({\color{blue} \bf +0.06}) & 14.70 ({\color{red} \bf -1.51}) & 10.79 ({\color{red} \bf -0.92}) & 14.90 ({\color{red} \bf -1.31}) & 11.83 ({\color{blue} \bf +0.12}) & 16.66 ({\color{blue} \bf +0.45}) \\
& Codegen-2B            & 20.91         & 27.75         & 18.48 ({\color{red} \bf -2.43}) & 26.56 ({\color{red} \bf -1.19}) & 17.87 ({\color{red} \bf -3.04}) & 26.13 ({\color{red} \bf -1.62}) & 22.50 ({\color{blue} \bf +1.59}) & 29.59 ({\color{blue} \bf +1.84}) \\
& Codegen-6B            & 24.02       & 36.82         & 26.71 ({\color{blue} \bf +1.69}) & 34.27 ({\color{red} \bf -2.55}) & 25.37  ({\color{blue} \bf +1.35}) & 34.02 ({\color{red} \bf -2.80})  & 25.73 ({\color{blue} \bf +1.71})  & 33.74  ({\color{red} \bf -3.08}) \\ \midrule
\multirow{5}{*}{MBPP} & Incoder-1.3B & 5.92  & 10.27 & 4.11 ({\color{red} \bf -1.82}) & 7.87 ({\color{red} \bf -2.40}) & 3.68 ({\color{red} \bf -2.25}) & 7.06 ({\color{red} \bf -3.21}) & 3.82 ({\color{red} \bf -2.10}) & 7.22 ({\color{red} \bf -3.05}) \\
& Incoder-6.7B & 7.53  & 11.55 & 7.75 ({\color{blue} \bf +0.23}) & 11.79 ({\color{blue} \bf +0.24}) & 7.86 ({\color{blue} \bf +0.34}) & 12.30 ({\color{blue} \bf +0.75}) & 7.80  ({\color{blue} \bf +0.28}) & 12.37 ({\color{blue} \bf +0.82}) \\
& Codegen-350M & 16.99 & 25.39 & 15.32 ({\color{red} \bf -1.67}) & 23.35 ({\color{red} \bf -2.04}) & 15.32 ({\color{red} \bf -1.67}) & 23.85 ({\color{red} \bf -1.54}) & 15.87 ({\color{red} \bf -1.12}) & 24.28 ({\color{red} \bf -1.12}) \\
& Codegen-2B   & 31.57 & 41.97 & 28.10 ({\color{red} \bf -3.47}) & 38.24 ({\color{red} \bf -3.73}) & 27.38 ({\color{red} \bf -4.19}) & 39.04 ({\color{red} \bf -2.93}) & 30.59 ({\color{red} \bf -0.98}) & 40.93 ({\color{red} \bf -1.04}) \\
& Codegen-6B   & 34.00 & 51.97 & 34.49 ({\color{blue} \bf +0.49}) & 45.42 ({\color{red} \bf -6.55}) & 34.74 ({\color{blue} \bf +0.49})  & 45.74 ({\color{red} \bf -6.23}) & 37.35 ({\color{blue} \bf +3.35})  & 48.90 ({\color{red} \bf -3.07}) \\  \bottomrule
\end{tabu}
} \vspace{-10pt}
\end{table*}

\noindent
\textit{Observations.}
{\bf CPU-based results.} In Table \ref{table:cpu_latency}, we report (on cloud and on device) hosting metrics of Codegen-2B/6B and Incoder-$6$B model for generating $20$ tokens for each example. As the quantization kernel in Pytorch only supports CPU inference, we collect all the metrics on CPUs. For both Codegen-6B and Incoder-6B, we observe that \texttt{int8} quantization reduces the model size to  about $29\%$ of \texttt{FP32} counterpart and also reduces latency significantly (e.g., by about 50\% on ec2 instance and $>60\%$ on laptop). As the carbon emission and pricing are roughly linear w.r.t. the runtime, using a quantized model would also contribute significantly to green AI and reduced hosting cost. With the much less stringent requirements on the underlying hardware, quantization makes it possible to run large (e.g., $6$B) code generation models on a personal laptop within a reasonable latency constraint. Such capability can be helpful for developers to get high-quality code recommendation/auto-completion in their local environment. 




{\bf GPU-based results.} In deep learning based generation models, typically the predominant portion of computational cost comes from the multiplication of various matrices. As Pytorch framework does not support GPU kernel-based end-to-end inference, we measure the potential latency impact through matrix multiplication as a proxy to showcase the efficacy of quantization. We report the latency results based on Nvidia CUTLASS kernel in Table \ref{table:gpu_latency}, and we can observe 40\% latency reduce across different sizes using \texttt{int8} matrix multiplication.

\RS{1}{The quantized models use much less latency, memory, and storage than the corresponding full precision model. It also has remarkably less carbon footprint. Thus, it is possible to fit even a 6B-parameter model within a regular laptop.}

\subsection{Accuracy Evaluation for Code Generation Task (RQ2)}

\noindent
\textit{Motivation.} Although greener, a quantized model will be mostly useful if it maintains the accuracy of the original full precision model. In this RQ, we evaluate the functional correctness of code generation models for full precision and their different quantized variants.


\noindent
\textit{Experimental Setup.} We evaluate the code generation tasks using CodeGen and Incoder quantized models with static and dynamic activation quantization. We tested the models with per-column scales and per-tensor scales while quantizing the weights as well. We report both pass@1 and pass@5 accuracies.

\begin{figure}
\vspace{-20pt}
     \centering
     \begin{subfigure}[b]{0.45\textwidth}
         \centering
         \includegraphics[width=0.8\linewidth]{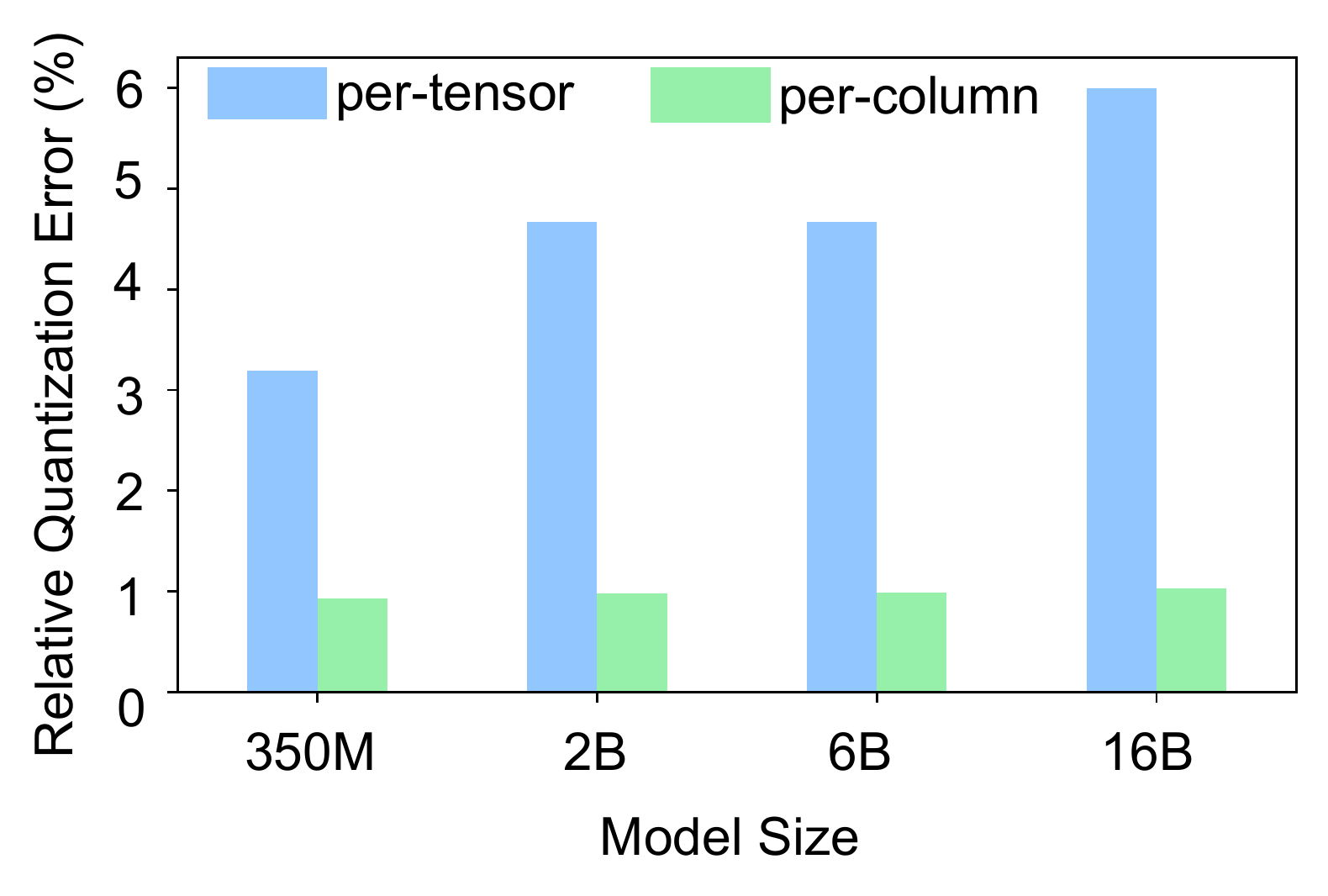}
         \vspace{-10pt}
         \caption{\textbf{\small{Weight quantization noise for CodeGen models across model sizes and granularity}}.}
         \label{fig:pert-v-perr}
     \end{subfigure}
     \hfill
     \centering
      \begin{subfigure}[b]{0.45\textwidth}
        \centering
         \includegraphics[width=0.85\textwidth]{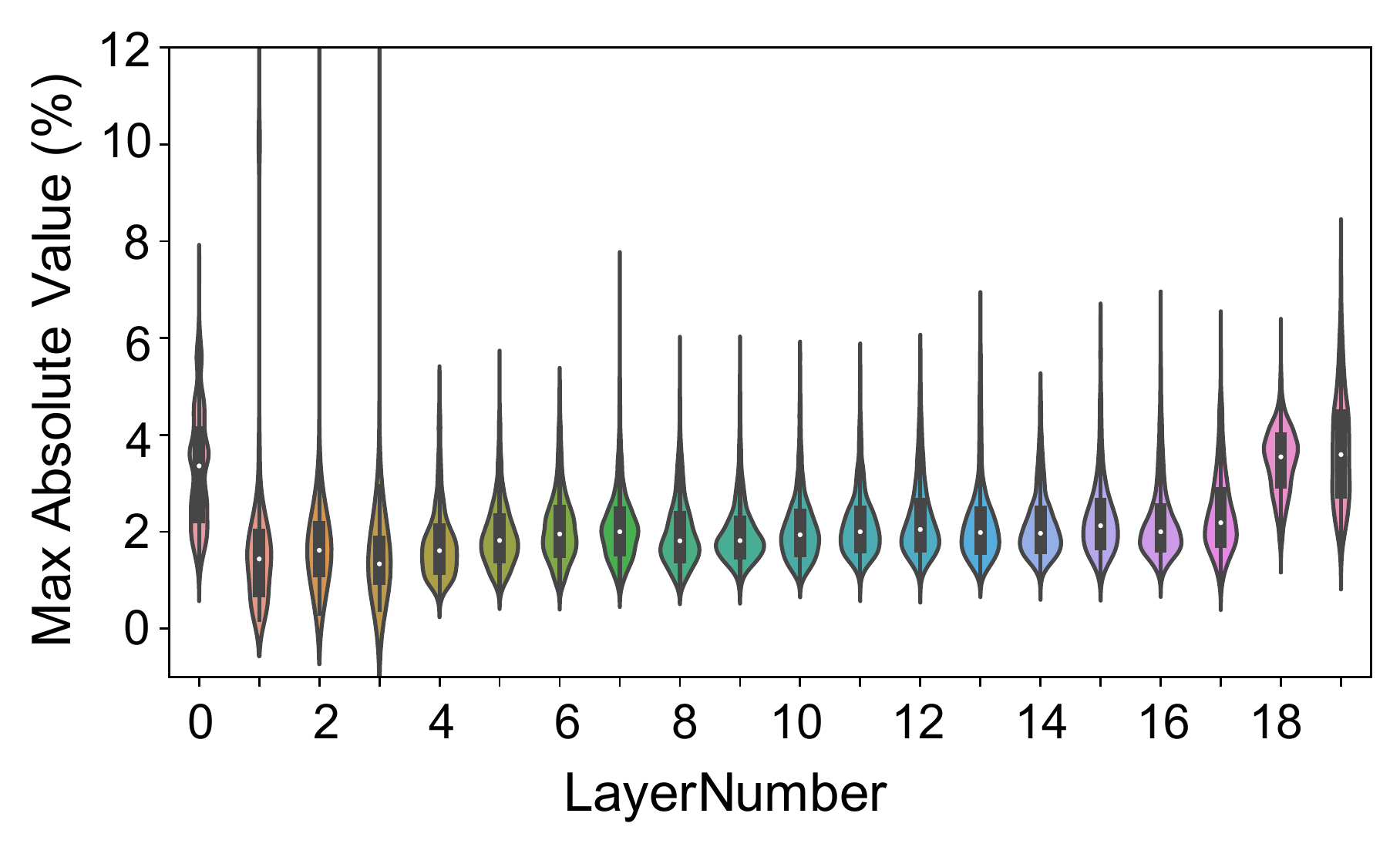}
         \vspace{-10pt}
         \caption{\textbf{\small Distributions of maximum activation value across different layers in CodeGen-350M model on validation data}.}
         \label{fig:dynamic}
     \end{subfigure}
     \hfill
     \begin{subfigure}[b]{0.45\textwidth}
         \centering
          \includegraphics[width=0.8\textwidth]{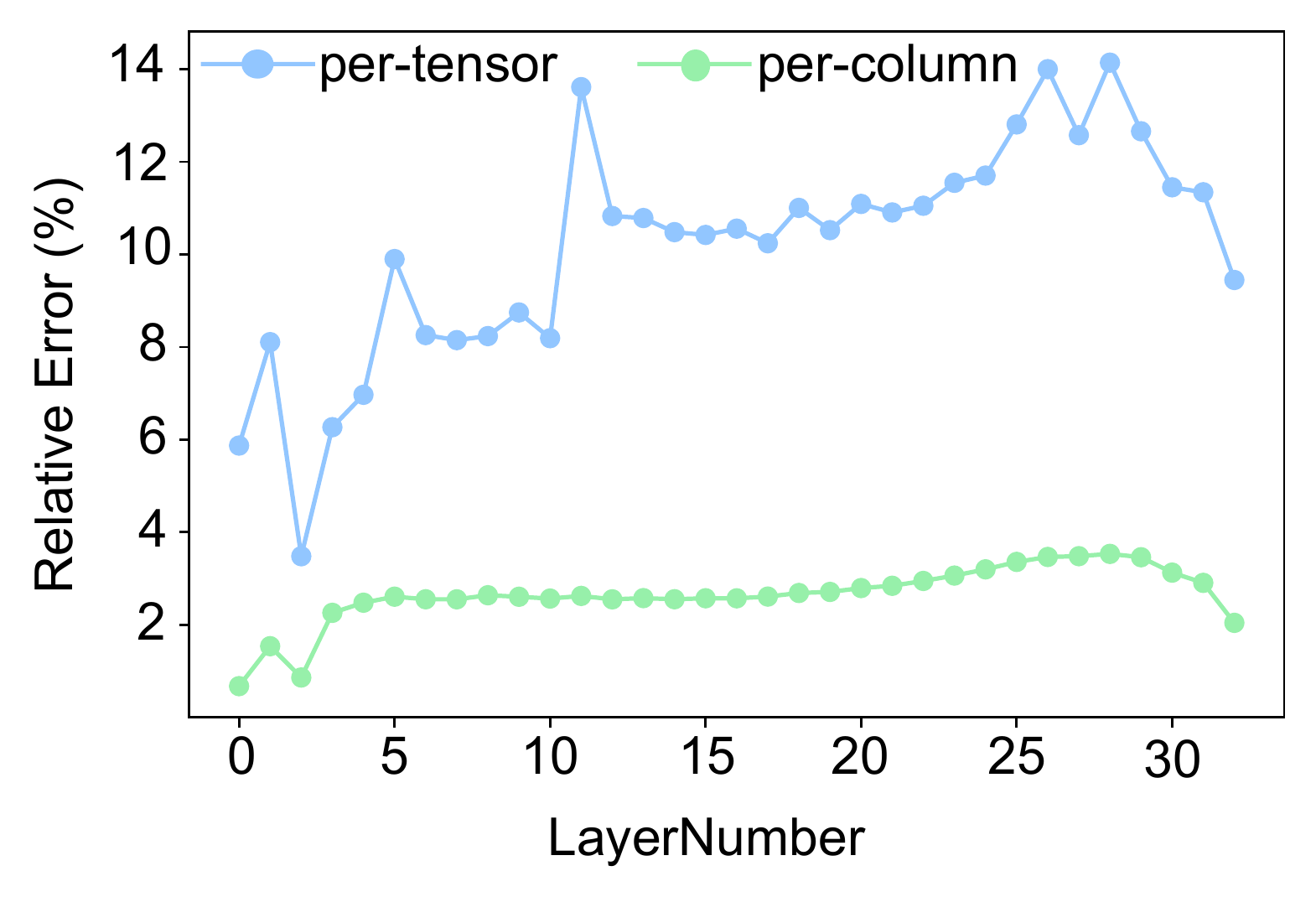}
          \vspace{-10pt}
         \caption{\textbf{\small{MSE/layer between activations from original vs.~quantized model. The error is estimated on a weight-only quantized CodeGen-6B model generating a token on HumanEval.}}}
         \label{fig:depth}
     \end{subfigure}
        \caption{\textbf{\small{Statistics impacting quantized model accuracy.}}}
        \label{fig:three graphs}
\vspace{-10pt}
\end{figure}

\noindent
\textit{Observations.} 
~\Cref{tab:rq2} summarizes the results.
We see accuracy gain for Incoder-6.7B models across all the quantization settings, while Incoder-1B shows an average accuracy drop of 0.84\% on HumanEval and 2.47\% on MBPP datasets.
CodeGen models show $<2\%$ average degradation with pass@$1$ metric on HumanEval and MBPP datasets with both Dynamic and Static quantization. However, we observe 3\%-4\% and 2\% average drop in accuracy in the pass@$5$ metrics with dynamic quantization and static (per-tensor) quantization respectively. 
With static (per-column) quantization the average pass@5 accuracy drop is  $<2\%$ for CodeGen models.

Overall, dynamic (per-tensor) quantization tends to outperform static (per-tensor) quantization by a small margin and static (per-column) quantization outperforms static (per-tensor) quantization. 
This is because:


\begin{itemize}[leftmargin=*]
    \item \textbf{Weight Quantization.}
Weight distributions have a high variance within a kernel, accounting for outliers that result in large quantization noise. This is particularly an issue with increasing matrix sizes in larger models. Figure~\ref{fig:pert-v-perr} shows how the quantization noise increases with model sizes with per-tensor scales, but not with per-column scales. This reduced quantization noise with per-column scales explains why static (per-column) setting outperforms static (per-tensor) one.

\item \textbf{Activation Quantization.} The primary challenge in activation quantization is in choosing the quantization scales. With static-quantization, we have to pick pre-determined scales based on validation data. This pre-determined scale is picked conservatively 
On the other hand, dynamic quantization allows us to adjust the scales for every input example and for every token, thereby making it attractive to reduce quantization noise. Dynamic quantization will be useful if we observe high variance in the max-values across different inputs/tokens. For example,~\Cref{fig:dynamic} shows the max value of the activation across different layers in CodeGen-350M.

\item  
\textbf{Error Accumulation.} Quantization noise accumulates with depth, making deeper models more challenging to quantization. ~\Cref{fig:depth} shows the relative quantization noise with model depth for CodeGen-6B model, showing quantization error growing with depth.  We observe that a) per-column quantization results in smaller accumulated error with depth and b) the error tends to reduce in the last few (\~4) layers of the model. The latter could be due to the inherent robustness of the model. 


\end{itemize}







\subsubsection{Ablation Study} 
To better understand the impact of different design choices on the model, as discussed in~\Cref{sec:prep}, we further investigated pass@$1$ scores 
for different model variations on HumanEval.

\begin{figure}[t]
     \centering
         \includegraphics[width=0.4\textwidth]{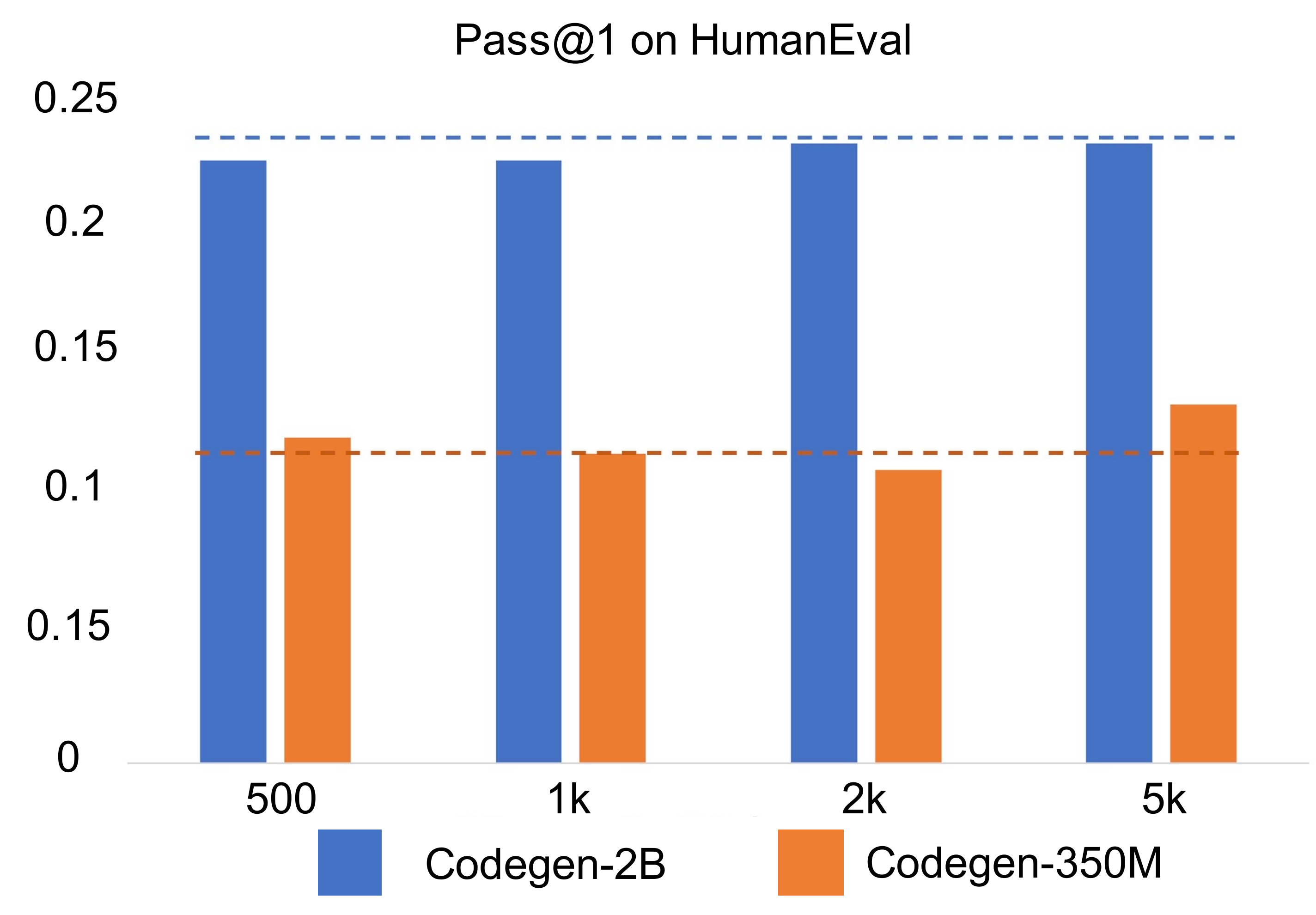}
         \caption{\textbf{\small{Execution accuracy on HumanEval with Codegen-2B and Codegen-350M (per-column static) when they are calibrated (\textit{MSE} loss) on different amounts of data (from $500$ to $5$k). Dotted lines denote the pass@$1$ of corresponding full-precision models}}.}
         \label{fig:calibaration_size}
\end{figure}

\noindent
\textbf{Size of calibration set.} Here, we study how the size of calibration data affects the performance of quantized models. \Cref{fig:calibaration_size} shows that the execution accuracy (on both 2B and 350M models) is typically stable across different sizes of calibration data. When using only $500$ samples for calibration, the quantized model can already learn a reasonable clipping range ($\alpha$) and achieve comparable accuracy as full-precision baselines. Such calibration cost (e.g., takes a few minutes on a single CPU/GPU) is almost negligible compared to other model compression options, such as distillation, which typically requires iterating over the whole training corpus and takes weeks to finish. 

\noindent
\textbf{Impact of precision.} We experimented with using 4-bit precision instead of the 8-bits that we use in the rest of the paper. The experiments with different precision settings on CodeGen-2B models on HumanEval and the results are summarized in Table~\ref{table:precision}. We use the static (per-column) quantization setting for these experiments.

With 8-bit weights and activation (W8A8), we can meet the accuracy of a full-precision model on HumanEval. However, this accuracy drops by $\approx 4\%$ with weights quantized with 4-bits while activations remain quantized with 8-bits (W8A4). We find that the model does not generate any meaningful outputs when activations are quantized with 4-bits while the weights remain quantized with 8-bits (W8A4), indicating that the model is more sensitive to activation quantization than those of the weights.

\begin{table}[h]
\caption{\small{\textbf{Execution accuracy of CodeGen-$2$B model at different activation and weight precision settings on HumanEval. Here WxAy indicates x-bit weights and y-bit activations.}}}
\label{table:precision}
\centering
\begin{tabular}{l|rr}
\toprule
\textbf{pass@} & \textbf{1} & \textbf{5} \\ \midrule
Full precision & 20.91\%    & 27.75\%    \\
W8A8           & 22.50\%    & 29.59\%    \\
W4A8           & 18.54\%    & 24.83\%    \\
W8A4           & 0.61\%     & 1.39\%     \\ \bottomrule
\end{tabular}
\end{table}

\subsubsection{Quantizing Extremely Large Code Generation Models}
\label{sec:extreme}

So far we have seen that appropriately designed quantization techniques 
could preserve accuracy for models with medium to large sizes (up to $6$B parameters).  Now we conduct an extreme study
with Codegen-$16$B, one of the largest publicly available code generation models. 

\begin{table}[t]
    \centering
\caption{\textbf{\small{Execution accuracy (pass@1 and pass@5) of Codegen-16B on HumanEval.}}}
\label{table:codegen_16B}
\begin{tabular}{l|rr}
\toprule
\textbf{pass@}      & \textbf{1} & \textbf{5} \\ \midrule
Full-precision      & 29.39\%    & 39.02\%    \\
Dynamic Quantization           & 27.68\%    & 39.63\%    \\
Static (per column) Quantization  & 26.40\%    & 34.78\%    \\ \bottomrule
\end{tabular}
\end{table}

From \Cref{table:codegen_16B}, one can observe that both dynamic and static (per-column) quantization achieve competitive results compared to the original model. For example, dynamic quantized model (model size: $17$ GB) could achieve similar pass@$5$ and slightly lower pass@$1$ compared to the significantly more gigantic \texttt{FP32} model ($75$ GB).

\RS{2}{Quantization models often suffer minimal accuracy drop from the corresponding full precision models making them potential design choices for implementing Greener Code Generation models.}

\subsection{Robustness Evaluation (RQ3)}


\begin{table}[t]
\centering
\caption{\small{\textbf{The percentage of the pass@1 {\em drop} on the datasets with character-level (Ch), word-level (W), and sentence-level (S) perturbations of prompt compared to the unperturbed ones.} We highlight the least drops for each setting.}} 
\label{table: robustness}
\setlength{\tabcolsep}{3pt}
\begin{tabular}{l|l|rrr|rrr}
\toprule 
\multicolumn{2}{c}{} & \multicolumn{3}{|c}{\bf HumanEval}                                                                              & \multicolumn{3}{|c}{\bf MBPP}                                                                                   \\ 
\multicolumn{2}{l}{} & \multicolumn{1}{|c}{\bf Ch} & \multicolumn{1}{c}{\bf W} & \multicolumn{1}{c|}{\bf S} & \multicolumn{1}{c}{\bf Ch} & \multicolumn{1}{c}{\bf W} & \multicolumn{1}{c}{\bf S} \\ \midrule
\multicolumn{8}{c}{\textbf{Incoder}} \\ \midrule

\multirow{4}{*}{\bf 1.3B}  & FP*          & {\color{red} \bf 0.00}                                 & 18.18                          & -9.09                             & 30.00                                & 35.00                          & 8.33                              \\
                               & D (T)       & 11.11                                & 11.11                          & 11.11                              & {\color{red} \bf 10.81}                                & 24.32                          & 13.51                              \\
                               & S (C) & {\color{red} \bf 0.00}                                 & 18.18                          &  0.00                               & 40.00                                & 30.00                          & {\color{red} \bf 7.50}                               \\
                               & S (T) & 10.00                                & {\color{red} \bf 10.00}                          & {\color{red} \bf -10.00}                             & 31.58                                & {\color{red} \bf  23.68}                          & 15.79                              \\ \midrule
\multirow{4}{*}{\bf 6.7B}  & FP          & {\color{red} \bf -7.69}                                & 30.77                          & 7.69                               & 24.68                                & 25.97                          & 10.39                              \\
                               & D (T)      & 7.69                                 & 7.69                           & 7.69                               & 18.42                                & 26.32                          & 15.79                              \\
                               & S (C) & 0.00                 & {\color{red} \bf 7.14}           & 14.29               & {\color{red} \bf 9.59}                 & {\color{red} \bf 19.18}           & {\color{red} \bf -4.11}               \\
                               & S (T) & -7.14                                & 14.29                          & {\color{red} \bf -7.14}                              & 25.97                                & 24.68                          & 7.79           \\ \midrule 
                               
\multicolumn{8}{c}{\textbf{Codegen}} \\ \midrule
                               
\multirow{4}{*}{\bf 350M}  & FP          & {\color{red} \bf 10.53}                                & {\color{red} \bf 10.53}                          & 15.79                              & 13.56                                & 19.21                          & 6.78                               \\
                               & D (T)      & 15.79                                & 15.79                          & {\color{red} \bf 5.26}                               & 17.72                                & 13.92                          & 7.59                               \\
                               & S (C) & 22.73                                & 18.18                          & 13.64                              & 14.91                                & {\color{red} \bf 12.42}                          & {\color{red} \bf 3.11}                               \\
                               & S (T) & 33.33                                & 23.81                          & 14.29                              & {\color{red} \bf 13.16}                                & 14.47                          & 5.26                               \\ \midrule
\multirow{4}{*}{\bf 2B}    & FP          & {\color{red} \bf 12.82}                                & {\color{red} \bf 15.38}                          & 20.51                              & 7.99                                 & {\color{red} \bf 9.27}                           & 6.39                               \\
                               & D (T))       & 29.73                                & 32.43                          & 27.03                              & 6.79                                 & 11.79                          & {\color{red} \bf -1.07}                              \\
                               & S (C) & 13.16                                & 23.68                          & 18.42                              & 10.03                                & 15.53                          & 7.12                               \\
                               & S (T) & 15.15                                & 27.27                          & {\color{red} \bf 12.12}                              & {\color{red} \bf 7.72}                                 & 9.56                           & 2.21                               \\ \midrule
\multirow{4}{*}{\bf 6B}    & FP          & 17.78                                & 24.44                          & 28.89                              & {\color{red} \bf -0.85}                                & {\color{red} \bf 4.55}                           & 0.28                               \\
                               & D (T)       & 30.00                                & 40.00                          & 34.00                              & 6.34                                 & 12.97                          & 6.05                               \\
                               & S (C) & 20.93                 & {\color{red} \bf 20.93}           & {\color{red} \bf 16.28}               & 6.96                 & 8.36           & {\color{red} \bf -0.84}               \\
                               & S (T) & {\color{red} \bf 15.56}                                & 28.89                          & 20.00                              & 6.10                 & 9.01           & 2.62               \\ \bottomrule
\end{tabular}

{\footnotesize *FP=Full-precision; D (T)=Dynamic (per-tensor); S(C)=Static (per-column); S(T)=Static (per-tensor)}
\end{table}

\begin{table*}[h]
\centering
\setlength{\columnsep}{0pt}
\caption{\small{\textbf{
Example impact of word-level, character-level, sentence-level perturbations on full-precision and per-tensor dynamic quantized models.} The perturbed region is \underline{underlined}.}}
\label{table: samples}

\resizebox{\textwidth}{!}{  
\begin{tabular}{l|l|l|c|c}
\toprule


\multicolumn{2}{c}{}                                                                               & \multicolumn{1}{|c}{}                                                                                & \multicolumn{2}{|c}{\bf Passing All Tests}                                   \\ 
\multicolumn{2}{c}{\multirow{-2}{*}{\bf Examples}}                                                     & \multicolumn{1}{|c}{\multirow{-2}{*}{\bf Docstring}}                                                     & \multicolumn{1}{|c|}{\bf Full-precision}                     & {\bf Dynamic (per-tensor)}               \\ \midrule

& Unperturbed     & Write a python function to determine whether all the numbers are different from each other are not. & \cmark                                & \cmark                                \\
& Character-level &  Write a python function to determine whet\underline{H}er al\underline{L} the numbers a\underline{R}e different from each othe\underline{R} are not.                         & \cmark                                & \cmark                                \\
& Word-level      & Write a python function to determine whether all the numbers are \underline{unlike} from each other are not.    & \cmark                                & \cmark                                \\
\multirow{-4}{*}{\bf S1} & Sentence-level   & \underline{Write a Python function to see if all numbers differ from each other.}  & \cmark                                & \cmark                                \\ \midrule

& Unperturbed     & Write a function to extract the index minimum value record from the given tuples.    & \cmark & \cmark                                \\
& Character-level & Write a function to extract the index mini\underline{M}um vaLue record fro\underline{M} the given tuples.                   & \cmark                                & {\color[HTML]{FF3633} \textbf{\xmark}} \\
& Word-level      & Write a function to extract the index \underline{minimal} value record from the give tuples.                    & \cmark                                & \cmark                                \\ 
\multirow{-4}{*}{\bf S2} & Sentence-level   & \underline{Write a function to extract the index minimum dataset from the given tuples.}                        & \cmark                                & \cmark                                \\ \midrule

& Unperturbed     & Write a function to print check if the triangle is equilateral or not.  & \cmark & \cmark                                \\
& Character-level & Write a function to print check if the tria\underline{N}gle i\underline{S} equilateral \underline{O}r not.  & \cmark & \cmark                                \\
& Word-level      & Write a function to print check if the triangle \underline{equal} equilateral or not.                           & {\color[HTML]{FF3633} \textbf{\xmark}} & \cmark                                \\
\multirow{-4}{*}{\bf S3} & Sentence-level   & \underline{Write a function to check whether the triangle is equilateral or not.}                               & {\color[HTML]{FF3633} \textbf{\xmark}} & \cmark                               \\ \bottomrule
\end{tabular}
}
\vspace{-10pt}
\end{table*}

\noindent
\textit{Motivation.} It is well known that Deep Learning models are sensitive to input perturbations~\cite{tian2018deeptest, han-etal-2020-adversarial, jin2020bert, zhang2020machine, ma2018deepgauge} ; i.e., a well-trained model performs significantly worse  when evaluated against meaningful perturbed inputs.  Thus, it is important to estimate {\em robustness} of a model by evaluating it against such perturbations.
In particular, a good quantized model should not adversely impact the robustness of a model, i.e., the original full-precision model's robustness should not decrease drastically after quantization. 

\noindent
\textit{Experimental Setup.} To evaluate the effect of quantization on a model's Robustness, we evaluate both the original and the quantized models on HumanEval~\cite{ChenM2021} and MBPP~\cite{AustinJ2021} dataset with perturbed inputs. In the NLP domain, researchers propose different semantic preserving perturbations to inputs; e.g., mutating words with their synonyms~\cite{ebrahimi2017hotflip,ren-etal-2019-generating,alzantot-etal-2018-generating} 
 or character-level mutations~\cite{8424632,zang-etal-2020-word}.
 We adapt similar techniques in our context. 
In particular, we perturb the text in each prompt with three different types of perturbations respectively (see~\Cref{table: samples}):
\begin{enumerate}[leftmargin=*]
    \item  \textit{Character-level perturbations} by changing randomly selected characters to upper cases.
    \item \textit{Word-level perturbations} by substituting randomly selected words with synonyms from WordNet~\cite{miller1995wordnet};
    \item \textit{Sentence-level perturbations} by paraphrasing the whole text with back translation~\cite{li2019improving,sugiyama2019data}. In specific, it transforms the English docstring into German and then translates back to English.
\end{enumerate}

For these three types of perturbations, we use the default settings and implementations from a standard text perturbation benchmark NL-Augmenter~\cite{dhole2021nl}. 
These perturbations are designed such that the original semantics of the natural language remains unaltered~\cite{goel2021robustness,zhang2022interpreting,morris2020textattack}.
Then we measure the average pass@1 with greedy sampling
for each model on the three perturbed datasets along with the unperturbed ones to avoid randomness and better observe the robustness trends.

To measure the robustness of a model, 
we compute the change in pass@1 results between perturbed and unperturbed inputs.
For each type of perturbation, we compute the percentage change across all the inputs in a dataset, as:  $\%\Delta = \frac{\mathrm{pass}@1_\mathrm{unperturbed}-\mathrm{pass}@1_\mathrm{perturbed}}{\mathrm{pass}@1_\mathrm{unperturbed}}$.


~\Cref{table: robustness} reports the results. The lower the value of $\Delta$, the better the robustness of a model. A negative drop means the model performs better with perturbed inputs. 

\noindent
\textit{Observations.}
The results show that, overall all the quantization methods, including per-tensor dynamic, per-tensor static, and per-column static, have comparable robustness performance w.r.t. the corresponding full precision model. 
In certain cases, in fact, quantized models perform better (as shown in red). 
On average across all model types and perturbations, full precision, per-tensor dynamic, per-tensor static, and per-column static quantized models have 13.27\%, 15.92\%, 12.91\%, and 13.33\% percentage of the drops on MBPP and HumanEval datasets. Models quantized with static per-column overall have slightly better robustness performance compared to the ones quantized with dynamic/static per-tensor quantized models. 

We further compute per sample difference in pass@1 result between a quantized and the 
corresponding full-precision model using Wilcoxon-Mann-Whitney test~\cite{fay2010wilcoxon}\textemdash this 
also confirms the difference between the two models in statistically insignificant.

\RS{3}{Quantization does not have any negative impact on model's robustness--- a quantized model reacts to perturbed inputs very similarly as the corresponding full-precision model.}


\subsection{Accuracy for Code Summarization Task (RQ4)}

\noindent
\textit{Motivation.} Here we check whether the quantization techniques studied so far are also applicable to other code-related tasks. In particular, we chose code summarization, as it is reversing the modality studied so far (NL for code).

\noindent
\textit{Experimental Setup.}
Here, we use the \emph{finetuned} PLBART and CodeT5 models on the code summarization task (in Python) released by the authors.
Since CodeGen is not designed to generate summaries given a code snippet, we do not use it in the evaluation. In our early experiments, we evaluated InCoder full precision models on this task based on the author released code, but got 
very poor performance, therefore, we do not pursue the model.

\noindent
{\bf Observations:} The results are presented in Table \ref{tab:summarization}. We observe almost no drop in BLEU score for CodeT5 models with both Dynamic and Static quantization. In comparison, while PLBART with Dynamic quantization matches the full-precision performance, we observe a performance drop with Static quantization. To understand this degradation in performance, we perform a qualitative comparison between these two settings. A few examples are provided in Table \ref{tab:qual_summarization}. Overall, we observe that PLBART with static quantization generates shorter summaries that affect the BLEU score. However, 
the generated summaries are semantically comparable to the full precision version.

\begin{table}[h]
\caption{\textbf{\small{Evaluation results (smoothed BLEU score) for code summarization.}}}
\label{tab:summarization}
\centering
\setlength{\tabcolsep}{1pt}
\begin{tabular}{l|cccc}
\toprule 
& {\bf Full-} & {\bf Dynamic } & {\bf Static } & {\bf Static } \\ 
& {\bf -precision} & {\bf (per-tensor)} & {\bf (per-tensor)} & {\bf  (per-column)} \\ 
\midrule
PLBART   & 17.02 & 17.00 ({\color{red} \bf -0.02}) & 14.96 ({\color{red} \bf -2.06}) & 15.19 ({\color{red} \bf -1.83}) \\
CodeT5   & 19.50 & 19.44 ({\color{red} \bf -0.06}) & 19.27 ({\color{red} \bf -0.23}) & 19.30 ({\color{red} \bf -0.20}) \\ \bottomrule
\end{tabular}
\end{table}

\begin{table}[t]
\centering
\caption{\small{\textbf{Qualitative comparisons of summaries by PLBART in full-precision and static quantization.}}}
\label{tab:qual_summarization}
\setlength{\tabcolsep}{2pt}
\begin{tabularx}{\columnwidth}{X|X}
\toprule
{\bf Full-precision} & {\bf Static (per-tensor)}  \\ \midrule
Copy an entire table to a temporary file.  & dump the contents of a table to a temporary file \\ \midrule
Recursively make all intermediate directories and subdirectories. & helper function to make intermediate dirs \\  \midrule
Downloads a video by its id and title. & download by vid \\ \midrule
Generate RST API documentation for a module. & Generate the documentation for the given module. \\
\bottomrule
\end{tabularx}
\end{table}

\RS{4}{Quantized models behave comparably to the corresponding full-precision models for code summarization tasks.}

\section{Threats to Validity}
\label{sec:threats}

This paper presents an in-depth empirical evaluation of a specific type of model compression technique on code generation task. The main threats to the validity of our conclusions are external, relating to the generalization of our findings, to both other types of compression techniques and to other ML-powered code related tasks. 

First, as discussed in~\Cref{section:background}, quantization-based compression techniques are mostly suitable for usecase as a typical developer may not have resources to retrain the model from scratch using other compression methods. \wasi{Why a developer would train an ML model? Can we rather say, "where we do not have resources"?}
\rayb{So Sec 2 says other compressioon techniques require full retraining.}

Second,  we focus on mostly generative tasks, and thus study code generation (NL-to-code) in detail. To evaluate the generalizability of our findings, we also investigate the effect of quantization on code summarization (RQ4). 

Finally, we have other threats including studying each of these tasks on two models and two dataset respectively.  However, these are state-of-the-art open source models and data widely studied in the literature. We further studied the different sizes of these models. We evaluated on perturbed data (RQ3) which also gives us confidence on the stability of our results. Besides, all the other quantization-related parameters used in the experiments are empirically evaluated. 
We also report the most stringent measurement (pass@1) 
to reduce any measurement bias.

\section{Conclusion} \label{conclusion}

Code Generation models based on large \plms 
have set the new state-of-the-art in generating functionally correct code given natural language description. However, the sizes of these models could be prohibitively large (e.g., billions of parameters), which can cause problems for green AI and responsible AI. Therefore, developing approaches towards improving model efficiency yet preserving their powerful generation capability is of great practical importance. In this paper, we address this problem by developing a quantization-based recipe for such models. 
We demonstrate the efficacy of proposed methods in terms of greenness, accuracy, and robustness. As future work, we would like to investigate the efficacy of quantization for more code intelligence applications, such as code search, code editing, and code translation.


\clearpage 
\small
\bibliographystyle{IEEEtran}
\bibliography{reference}
\end{document}